\documentclass{article}
\usepackage{graphicx} 
\usepackage{tabularx}
\usepackage{booktabs}
\usepackage[utf8]{inputenc}
\usepackage{hyperref}
\usepackage{amssymb}
\usepackage{float}
\usepackage{caption}
\usepackage{authblk}
\usepackage{titling}
\title{AutomationBench}
  \author{Daniel Shepard}
  \author{Robin Salimans}
  \affil{\texttt{daniel.shepard@zapier.com}}
  \affil{\texttt{robin.salimans@zapier.com}}
\date{April 2026}

\begin{document}

\maketitle
\begin{abstract}Existing AI benchmarks for software automation rarely combine cross-application coordination, autonomous API discovery, and policy adherence. Real business workflows demand all three: a single task may span a CRM, inbox, calendar, and messaging platform — requiring the agent to find the right endpoints, follow a policy document, and write correct data to each system. To address this gap, we introduce AutomationBench, a benchmark for evaluating AI agents on cross-application workflow orchestration via REST APIs. Drawing on real workflow patterns from Zapier's platform, tasks span Sales, Marketing, Operations, Support, Finance, and HR domains. Agents must discover relevant endpoints themselves, follow layered business rules, and navigate environments with irrelevant and sometimes misleading records. Grading is programmatic and end-state only: whether the correct data ended up in the right systems. Even the best frontier models currently score below 10\%. AutomationBench provides a challenging, realistic measure of where current models stand relative to the agentic capabilities businesses actually need.\end{abstract}

\begin{figure}[H]
    \centering
    \includegraphics[width=0.86\linewidth, keepaspectratio]{Diagram.png}
    \caption{AutomationBench evaluation pipeline. The agent iteratively discovers and calls API endpoints to complete a task. Only the final state of the simulated applications is graded.}
    \label{fig:pipeline}
\end{figure}

\section{Introduction}
Businesses run on a variety of interconnected web applications: CRMs, email, calendars, spreadsheets,
project management tools, marketing platforms, support systems, messaging services, and more.
Coordinating actions across these applications is one of the most time-consuming aspects of
knowledge work, requiring agents to follow organizational policies, reason about which
information to act on, and ignore irrelevant or misleading context.

AI agents promise to automate these workflows, but existing benchmarks measure adjacent
capabilities rather than cross-application business orchestration. WebArena~[1] and
Mind2Web~[2] evaluate long-horizon web tasks in browsing environments, and OSWorld~[3]
evaluates open-ended computer-use tasks spanning real web and desktop applications. Tool
and service-workflow benchmarks evaluate tool/API use (and, in some cases, end-state
success) under constrained settings, for example, fixed tool corpora with retrieval-based
selection (ToolBench~[4], API-Bank~[5]), or a small number of simulated apps/domains with
programmatic evaluation (AppWorld~[6], $\tau^3$-bench~[7,8,9]). While $\tau^3$-bench recently added doc-based tool discovery in its banking domain, each task still operates within a single application rather than requiring cross-application coordination.

  \begin{table}[ht]
    \centering
    \footnotesize
    \resizebox{\textwidth}{!}{%
    \begin{tabular}{@{} l c c c c c l @{}}
    \toprule
    \textbf{Benchmark} & \textbf{Domains / Apps} & \textbf{API Discovery} & \textbf{End-State Grading} & \textbf{Cross-App} & \textbf{Business Rules} & \textbf{Environment} \\
    \midrule
    WebArena~[1]        & 4 websites     & \texttimes & \checkmark & Limited & \texttimes & Containerized web apps \\
    Mind2Web~[2]        & 137 websites   & \texttimes & \texttimes & \texttimes & \texttimes & Snapshots \\
    OSWorld~[3]         & Real OS + apps & \texttimes & \checkmark & \checkmark & \texttimes & Real OS \\
    ToolBench~[4]       & 16k+ APIs      & Retrieval  & \texttimes & \texttimes & \texttimes & RapidAPI \\
    API-Bank~[5]        & 73 APIs        & Retrieval  & \texttimes & \texttimes & \texttimes & Simulated \\
    AppWorld~[6]        & 9 apps         & Given      & \checkmark & \checkmark & \texttimes & Simulated \\
      $\tau^3$-bench~[7,8,9] & 4 domains   & Partial & \checkmark & \texttimes & \checkmark & Simulated \\

    \midrule
    \textbf{AutomationBench} & \textbf{6 domains/47 apps} & \textbf{Retrieval} & \textbf{\checkmark} & \textbf{\checkmark} & \textbf{\checkmark} & \textbf{Simulated} \\
    \bottomrule
    \end{tabular}}
    \caption{Comparison of AutomationBench with existing benchmarks. Units in the \textit{Domains / Apps} column vary by benchmark to reflect how each defines its scope. \textit{API Discovery} indicates whether the agent must find relevant endpoints itself. \textit{Cross-App} indicates whether tasks require coordinating actions across multiple independent applications. \textit{Business Rules} indicates whether tasks embed organizational policies that the agent must consult and follow.}
    \label{tab:benchmark-comparison}
  \end{table}

AutomationBench fills this gap. Agents receive a natural language business task and must
autonomously discover available REST API endpoints across dozens of integrated applications,
make sequential and interdependent calls, apply layered business rules, and navigate
environments seeded with distractors, decoy records, and adversarial inputs. Tasks span
Sales, Marketing, Operations, Support, Finance, and HR domains. Grading is purely on end-state
correctness, whether the right data ended up in the right systems, using programmatic
assertions against a simulated world state. This reflects how businesses actually evaluate
automation.

Our objective is to accelerate research on cross-application agent orchestration — enabling AI agents to coordinate actions across diverse web applications.

Though solutions like MCPs and CLI tools exist, REST APIs are the universal standard and frequently the glue behind those tools. Direct, accurate use of REST APIs dramatically expands the set of applications agents can interact with without requiring purpose-built integrations. Computer use offers broader access but at significantly higher computational cost, making API-based orchestration a more practical near-term path for high-volume business automation.

Advances in multi-step reasoning, policy adherence, and tool discovery would unlock a new class of complex automation tasks that current models cannot reliably complete. AutomationBench is designed to expose and measure this gap, directing research effort toward the capabilities businesses need most.

\section{Dataset construction and validation}
\begin{table}[h]
  \centering
  \begin{tabular}{lcc}
  \toprule
  \textbf{Domain} & \textbf{Public} & \textbf{Private (Eval)} \\
  \midrule
  Sales       & 100 & 100+ \\
  Marketing   & 100 & 100+ \\
  Operations  & 100 & 100+ \\
  Support     & 100 & 100+ \\
  Finance     & 100 & 100+ \\
  HR          & 100 & 100+ \\
  \midrule
  \textbf{Total} & \textbf{600} & \textbf{600+} \\
  \bottomrule
  \end{tabular}
  \caption{Task distribution across domains. Public tasks are used for research and experimentation; private tasks are held out exclusively for evaluation.}
  \label{tab:task-distribution}
  \end{table}

We created realistic tasks for six distinct domains: Sales, Marketing, Operations, Support, Finance, and HR. These domains are the most popular types of workflows Zapier customers use to automate their businesses. A pool of $\sim$500 API endpoints across 47 apps is used across the different domains.

To maintain the integrity of the benchmark, at least half of the tasks are private. Currently, there are 100 public tasks per domain and even more private ones. Scores will be posted to the leaderboard based on the private dataset. The private dataset started from a similar distribution as the public but has some additional hardening techniques applied.

Tasks were synthetically generated based on use cases from real customers. No PII or confidential customer data was used — only the shape of workflows sent along with negative feedback on Zapier's Agents service. We were targeting workflows that are difficult to set up. We clustered these and used the relevant entries for each domain. We fed these workflow patterns into models for inspiration and made many passes on realism, difficulty, and variety of apps. We used Opus 4.6, GPT 5.3 Codex, and Gemini 3 in task generation. To encourage growth in reasoning and be a challenging benchmark, we hardened tasks that were too easy. Few tasks should be easy enough to be solvable by the smallest models. In tasks, we try to capture the complexity of real operations by multiple layers of difficulty and noise.

We used a variety of hardening techniques, including, but not limited to, adding irrelevant data, keeping key info behind tool call responses, ambiguity of where info can be found, similar naming for incorrect entries, and strict business policy rules with overriding priorities. We developed these techniques through experiments on what is challenging for current models. The private dataset employs additional hardening techniques to prevent overfitting on a small set of challenges. Benchmark progress ideally reflects improved reasoning over novel complexity rather than mastery of a few known difficulty types.

Many business workflows require subjective judgment — assessing lead quality, determining urgency, or choosing a routing path. We limit subjectivity in this benchmark, but some degree is supported in these tasks by requiring the agent to record its decision as a structured state change (e.g., updating a status field, assigning a score, routing to a queue) rather than producing free-form text. Assertions then verify the recorded decision and any required downstream effects.

Some tasks require the agent to follow policy documents that override seemingly reasonable defaults or external requests. Although this tests an important real-world skill (adhering to organizational rules over intuition), it superficially resembles a prompt injection. To avoid conflating policy-following with prompt-injection compliance, task prompts explicitly reference these policy documents, signaling that consulting them is expected behavior rather than an adversarial trick.

Every task must satisfy all of the following:
\begin{itemize}
  \item \textbf{Observable outcome:} The task must produce at least one concrete state mutation (object created, field updated, label applied, event scheduled, activity logged, etc.).
  \item \textbf{Deterministic validation:} Correctness must be decidable using invariant-based checks, not subjective interpretation.
  \item \textbf{No free-form success criteria:} Success must be expressible as structured invariants over the world state.
  \item \textbf{Reasoning must commit:} If a task requires judgment or reasoning, the agent must record the decision in the system state.
  \item \textbf{Autonomous:} A single instruction must be given. Then the task is non-interactive until completion.
\end{itemize}

Example task (Full examples are in the Appendix):

\begin{quote} 
There's a scheduling conflict on February 20, 2026 at 2:00\,PM --- a Zoom meeting
and a Google Calendar event overlap. Check the meeting priority policy in the
spreadsheet to determine which one wins, then reschedule the loser by prepending
\texttt{[RESCHEDULED]\textvisiblespace} to its topic/title. Post a summary to
\texttt{\#ops-updates} on Slack noting which meeting won and which was rescheduled,
including both the Zoom meeting ID and Calendar event ID.
\end{quote}

To verify that tasks are solvable, we created a separate hint sheet that provides additional guidance for each task within a separate runtime mode. These include instructions on which endpoints to call and clues on parameters but not the parameters themselves. This allowed even smaller models to score around 80–100\% (such as Haiku). All tasks should be expected to achieve 100\% accuracy with larger models (such as Opus), though single-digit variance still persists.  We inspected a sampling of tasks and repeated audits to catch errors.

Hints deliberately exclude actual parameter values, as including them could bypass tool calls entirely and mask discoverability issues. We audited hints against a rules document to catch such violations.

Because the benchmark was built on Prime Intellect's Verifiers framework, we were able to pressure-test the benchmark by running Reinforcement Learning with Verifiable Rewards experiments with minimal setup in their Lab. These runs identified some common reward-hacking strategies that we used to strengthen the reward function. 

\section{Tool interface}
Public API schemas are provided for each app. To access these, the agent must use two tools: Search and Execute. Search performs keyword searches (BM25) across all available API schemas (top-k of 5). Execute mimics a curl or fetch request accepting method, url, and body. No authentication is mimicked. Finding the right endpoints to use is part of the challenge. The correct app or action name might not be known. The agent must explore the available tools and figure it out. The execute tool triggers a simulation of retrieving or updating a database through Pydantic Models. These Pydantic models serve as the source of truth for the state of the app. 
We preserve the \textit{shape} of real API schemas, including pagination, required fields, and common error cases (4xx status codes). 

We specify a maximum of 50 steps. This limit is rarely hit (tasks in the single digits). We can increase it if future models regularly hit this limit. Multiple tools can be called in parallel per step.

\section{Scoring}
For this benchmark, instead of grading a text response or document, we consider only the end state of each connected app. How the Agent got there is not a concern (what tools were called in what order or way), only the end state. Sometimes, multiple API endpoints can produce similar results. So, if the agent was supposed to send an email, create a lead, and update another lead, the grader checks the “sent mailbox folder” and the lead entries and verifies that all expected data is correct. If an agent takes multiple actions inefficiently (create the lead, update the lead 5 times instead of once), it yields the same end result and scores equally, but with a higher Cost.

Focusing on end-to-end results gives the agent flexibility to solve problems in creative ways (such as parallel tool calls and batched changes) while avoiding being overly prescriptive. It also keeps the tool's side effects in focus as that is what businesses care about.

The assertions for each task are used to deterministically verify the state to define success. We give no partial credit but we can see how close models get to completion for non-scoring purposes.
Businesses do not want partially completed workflows. The task must be completed to earn a positive score. Scores can then be compared across models.

Scores are for a single run. Run-to-run variance is typically within 1\%.

We also report Cost (USD per run) alongside Score on the leaderboard,
acknowledging that two models with the same pass rate may differ substantially
in efficiency. Cost is computed by taking the average token cost spent per task. Time is harder to compare across models and providers due to rate limiting (especially on unreleased alpha models with limited compute). So Cost is the primary efficiency metric at this point.

We include negative assertions to prevent shotgun approach reward hacking (sending the email multiple times to everyone in the company instead of just the specified recipient).

We use deterministic assertions instead of LLM-as-a-judge. All success
criteria are expressed as exact string matches and structural checks over the
world state, without semantic or subjective evaluation of text. This ensures
scores are largely reproducible across runs and eliminates some potential inter-model grading
bias. We want to minimize the score of one model based on another model's judgement (though synthetically generated data still carries some of this risk).

\section{Limitations}
As with any synthetically generated data, there is a risk of lack of realism and impossibility. The scale and complexity of tasks also make it challenging to manually review all tasks in a reasonable time frame. We manually inspected samples of tasks to reduce this risk.

We used various hardening techniques to realistically increase the difficulty of the tasks. Once these challenges are overcome, there will likely still be some reasoning gaps that matter for real business workflows. We can address these with later versions of the benchmark.

We regularly audited for bugs and unrealistic complexity, but there is still room for improvements. We plan to continue fixing bugs and will version major changes.
\section{Leaderboard}

\begin{table}[H]
\centering

\begin{tabular}{ l  l  ll }
\toprule
\textbf{Model} & \textbf{Score} &  \textbf{Cost per task}\\
\midrule

Opus 4.7 (max) & 9.9\% &  \$1.80 \\

Gemini 3.1 Pro (high) & 9.6\% &  \$0.54 \\

GPT 5.4 (high) & 7.6\% &  \$1.93 \\

Sonnet 4.6 (max) & 5.3\% &  \$1.81 \\

Haiku 4.5 & 1.5\% &  \$0.18 \\

GPT 5.4 (no reasoning) & 1.2\% &  \$0.19 \\
\bottomrule

\end{tabular}

\end{table}

\begin{figure}[H]
    \centering
    \includegraphics[width=1\linewidth]{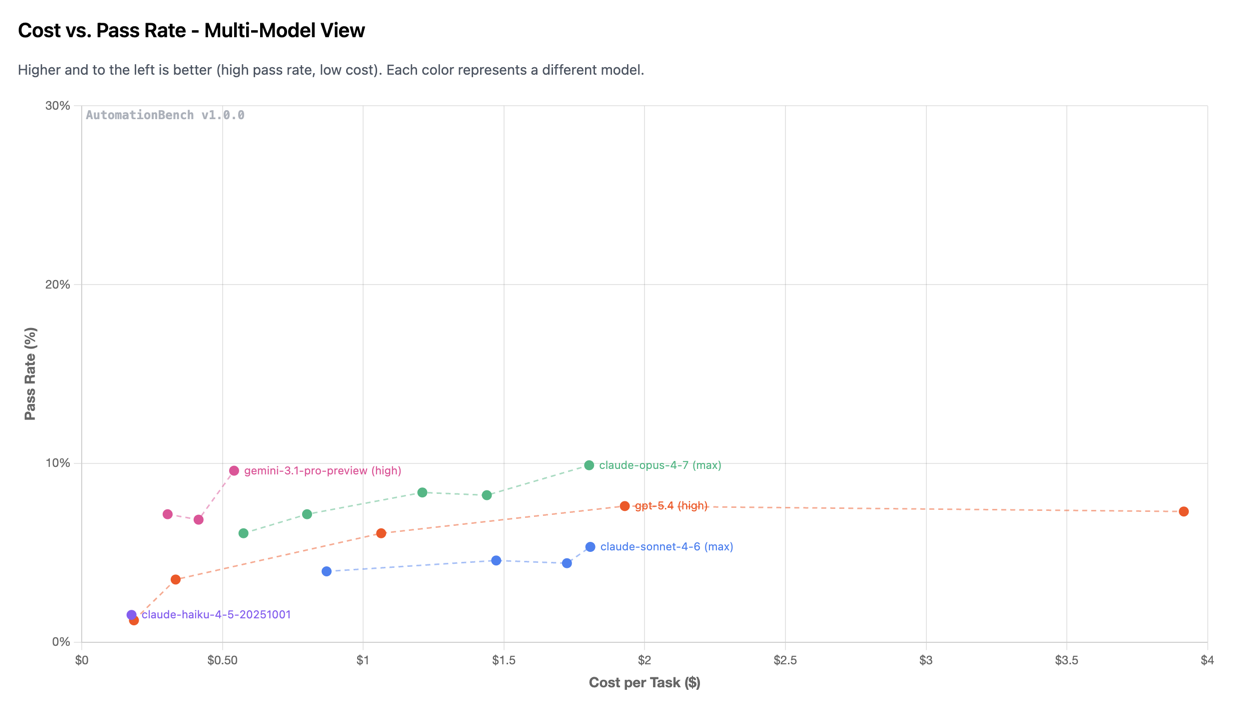}
    \caption{Model Results with Reasoning Effort}
    \label{fig:leaderboard}
\end{figure}

\section{Results}
State-of-the-art models all score below 10\%. Opus 4.7 tops the leaderboard at 9.9\%. See our benchmark leaderboard page~[10] for up to date results.

Models are solving fairly different sets of tasks from each other. Gemini and Opus are the top scoring models. Their passing sets have
  a Jaccard similarity of 0.17. 71\% of the tasks solved by Opus are not solved
  by Gemini, and 70\% of Gemini's passes are missed by Opus.

Sonnet 4.6 generates so many more tokens than Opus 4.7 that it costs more per task at every reasoning effort level, while also achieving a lower pass rate. Gemini achieves some of the highest accuracy at a fraction of the cost.

Opus uses an average of 12.6 steps and with 29.8 tool calls per task whereas Gemini uses 21.8 steps and 35.4 tool calls. GPT 5.4 high uses a similar number of steps at 15.4 but uses more tool calls at 43.9.

Given the multiple layers of complexity, we created some different toolsets and modes to diagnose issues and model deficiencies.

We built an alternative Zapier toolset using tool schemas based on Zapier's internal actions. This toolset usually results in an increased score over the API toolset, probably because they are more digestible than raw API schemas.

There is also a Limited Zapier toolset that uses only the specific Zapier tools necessary to perform the task instead of the typical Search + Execute tools. Some models benefit disproportionately from a narrower tool surface.

Gemini 3.1 Pro passes 9.6\% (API/default), 12.8\% (Zapier), and 14.3\%
  (Limited Zapier); Haiku 4.5 passes 1.5\%, 2.0\%, and 3.8\% respectively.

We also created a simple domain as a baseline to signal how well models do with simpler workflows without the noise and difficulty of regular tasks. This is not used in the benchmark score, but as a test of the harness. Even Haiku scored 97\% on this domain. For simple tasks with explicit instructions, today's models do just fine.

All of these toolsets and public tasks are available to try out from our public repository~[11] and Prime Intellect's Environments Hub~[12].

Having multiple failure modes and layers of complexity requires model reasoning to grow more capable in multiple ways rather than conquering it with a single improvement. This makes the benchmark more resilient and relevant as models improve.

\subsection*{Common failure modes}
More often than not, models declared success while actually failing. 72\% of Opus's failures, 91\% of Gemini's, and 84\% of GPT 5.4's involved this false confidence. One common failure mode for completing tasks is not finding data. Models often are not persistent or methodical enough in searching for data when generic searches do not find the results. They make assumptions about where data should live — e.g., assuming a CRM instead of Google Sheets, when that is not always the case. Another failure mode is missing items. When a task involves processing a list (12 emails, 8 leads), models frequently process some items correctly and then summarize as if done without verifying every item was covered. Models also do not follow instructions exactly. They paraphrase or outright ignore requirements for the content to include certain values despite precise instructions being given.

\section{Acknowledgements}
I (Daniel) want to thank Robin Salimans for helping create the architecture and offering technical guidance. We also thank Jake Talgard for advice and assistance with stakeholder communication. Thanks to Anna Marie Clifton, our director of product, for helping kickstart and prioritize this benchmark. Mike Knoop provided valuable feedback and expertise on benchmarks with ARC Prize. We thank Jichao Yin for brainstorming and sanity checking. Thanks to Prime Intellect, Anthropic, and OpenAI for their early feedback, and to many others at Zapier who cleared the way and supported us to focus on building this.

\section{Disclaimer}

The company names, product names, and API schemas referenced in this benchmark are trademarks or intellectual property of their respective owners. API schemas are based on publicly available documentation and are used solely for research and evaluation purposes under fair use. App behavior is simulated through synthetic models and does not represent exact production behavior. All data within tasks is entirely fictional — no real user data, transactions, or confidential information from any third party is included. This benchmark is not affiliated with, endorsed by, or sponsored by any of the companies whose products are referenced.

\section{References}
\begin{enumerate}
  \item S. Zhou, F. F. Xu, H. Zhu, X. Zhou, R. Lo, A. Sridhar, X. Cheng, T. Ou,
        Y. Bisk, D. Fried, U. Alon, G. Neubig.
        \textit{WebArena: A Realistic Web Environment for Building Autonomous Agents.}
        ICLR, 2024.

  \item X. Deng, Y. Gu, B. Zheng, S. Chen, S. Stevens, B. Wang, H. Sun, Y. Su.
        \textit{Mind2Web: Towards a Generalist Agent for the Web.}
        NeurIPS, 2023.

  \item T. Xie, D. Zhang, J. Chen, X. Li, S. Zhao, R. Cao, T. J. Hua, Z. Cheng,
        D. Shin, F. Lei, Y. Liu, Y. Xu, S. Zhou, S. Savarese, C. Xiong, V. Zhong, T. Yu.
        \textit{OSWorld: Benchmarking Multimodal Agents for Open-Ended Tasks in Real
        Computer Environments.}
        NeurIPS, 2024.

  \item Y. Qin, S. Liang, Y. Ye, K. Zhu, L. Yan, Y. Lu, Y. Lin, X. Cong, X. Tang,
        B. Qian, S. Zhao, L. Hong, R. Tian, R. Xie, J. Zhou, M. Gerstein, D. Li,
        Z. Liu, M. Sun.
        \textit{ToolLLM: Facilitating Large Language Models to Master 16000+ Real-World
        APIs.}
        ICLR, 2024.

  \item M. Li, Y. Zhao, B. Yu, F. Song, H. Li, H. Yu, Z. Li, F. Huang, Y. Li.
        \textit{API-Bank: A Comprehensive Benchmark for Tool-Augmented LLMs.}
        EMNLP, 2023.

  \item H. Trivedi, T. Khot, M. Hartmann, R. Manku, V. Dong, E. Li, S. Gupta,
        A. Sabharwal, N. Balasubramanian.
        \textit{AppWorld: A Controllable World of Apps and People for Benchmarking
        Interactive Coding Agents.}
        ACL, 2024.

  \item S. Yao, N. Shinn, P. Razavi, K. Narasimhan.
      \textit{$\tau$-bench: A Benchmark for Tool-Agent-User Interaction in Real-World
      Domains.}
      arXiv preprint arXiv:2406.12045, 2024.
\item V. Barres, H. Dong, S. Ray, X. Si, K. Narasimhan.
      \textit{$\tau^2$-Bench: Evaluating Conversational Agents in a Dual-Control
      Environment.}
      arXiv preprint arXiv:2506.07982, 2025.
\item Q. Shi, A. Zytek, P. Razavi, K. Narasimhan, V. Barres.
      \textit{$\tau$-Knowledge: Evaluating Conversational Agents over Unstructured
      Knowledge.}
      arXiv preprint arXiv:2603.04370, 2026.

  \item AutomationBench Leaderboard. \url{https://zapier.com/benchmarks}

  \item AutomationBench Public Repository. \url{https://github.com/zapier/AutomationBench}

  \item Prime Intellect Environment. \url{https://app.primeintellect.ai/dashboard/environments/zapier/AutomationBench}
\end{enumerate}

\section{Appendix: Task Examples}

Example task data is simplified and includes additional notes for illustration purposes.

 \subsection*{Task 1 · Sales — Meeting Conflict Resolution}
 
\subsubsection*{Prompt}
\begin{quote}
There's a scheduling conflict on February 20, 2026 at 2:00\,PM --- a Zoom meeting
and a Google Calendar event overlap. Check the meeting priority policy in the
spreadsheet to determine which one wins, then reschedule the loser by prepending
\texttt{[RESCHEDULED]\textvisiblespace} to its topic/title. Post a summary to
\texttt{\#ops-updates} on Slack noting which meeting won and which was rescheduled,
including both the Zoom meeting ID and Calendar event ID.
\end{quote}
 
\subsubsection*{Relevant API Endpoints}
 
{\footnotesize
\begin{tabular}{@{} l l l @{}}
\toprule
\textbf{Method} & \textbf{Endpoint} & \textbf{Purpose} \\
\midrule
GET   & \texttt{<zoom>/v2/users/me/meetings}                      & List Zoom meetings \\
GET   & \texttt{<calendar>/v3/users/me/calendarList}              & Find calendar IDs \\
GET   & \texttt{<calendar>/v3/calendars/\{id\}/events}            & List events on Feb 20 \\
GET   & \texttt{<gmail>/v1/users/\{userId\}/messages}             & Check inbox for policy override emails \\
GET   & \texttt{<sheets>/v4/spreadsheets/\{id\}/values/\{range\}} & Read meeting priority policy \\
PATCH & \texttt{<zoom>/v2/meetings/\{meetingId\}}                 & Update losing meeting's topic \\
POST  & \texttt{<slack>/chat.postMessage}                         & Notify \texttt{\#ops-updates} \\
\bottomrule
\end{tabular}}
 
\subsubsection*{Environment Setup}
 
\paragraph{Zoom --- 4 meetings.}
The conflicting meeting is highlighted in bold.
 
{\footnotesize
\begin{tabularx}{\textwidth}{@{} l l l X @{}}
\toprule
\textbf{ID} & \textbf{Topic} & \textbf{Start (UTC)} & \textbf{Notes} \\
\midrule
\textbf{1234567890}
  & \textbf{Q1 Product Review -- External}
  & \textbf{14:00}
  & 2 external registrants; agenda: ``Quarterly product review with partners'' \\[4pt]
1111111111
  & Product Review -- Internal
  & 10:00
  & Agenda: ``PRIORITY: Must attend'' \\[4pt]
9988776655
  & Partner Onboarding -- PRIORITY
  & 15:00
  & Agenda: ``PRIORITY: Onboarding session for new external partners'' \\[4pt]
5544332211
  & \texttt{[RESCHEDULED]} Q4 Pipeline Review
  & 14:00
  & Already prefixed; internal carry-over \\
\bottomrule
\end{tabularx}}
 
\bigskip
\paragraph{Google Calendar --- \texttt{primary} calendar, 3 events.}
The conflicting event is highlighted in bold.
 
{\footnotesize
\begin{tabularx}{\textwidth}{@{} l l l X @{}}
\toprule
\textbf{ID} & \textbf{Summary} & \textbf{Time (UTC)} & \textbf{Notes} \\
\midrule
\textbf{evt\_conflict\_001}
  & \textbf{Executive Strategy Session}
  & \textbf{14:00--15:00}
  & Attendees: \texttt{cto@\ldots}, \texttt{manager@\ldots};
    description: ``Strategic planning with CTO'' \\[4pt]
evt\_other\_001
  & Team Standup
  & 09:00--09:30
  & Attendees: \texttt{team@\ldots} \\[4pt]
evt\_tz\_trap\_001
  & APAC Sales Sync
  & 19:00--20:00
  & Description mentions ``2:00\,PM PST'' but actual start is 19:00\,UTC \\
\bottomrule
\end{tabularx}}
 
\bigskip
\paragraph{Google Sheets}
(\texttt{ss\_meeting\_policy}\,/\,\texttt{ws\_priority\_rules}) ---
priority rules, highest first:
 
\begin{enumerate}
  \item \textbf{Rule 1 (highest):} Meeting has a C-level attendee
        (CTO, CEO, CFO, COO, CIO).
  \item \textbf{Rule 2:} Meeting has \texttt{PRIORITY} in agenda/description.
  \item \textbf{Rule 3:} Meeting includes external attendees.
  \item \textbf{Rule 4 (lowest):} Internal meeting with no special flags.
\end{enumerate}
 
\noindent\fbox{\parbox{\dimexpr\textwidth-2\fboxsep-2\fboxrule}{%
\textbf{Policy note:} Only internal leadership emails may amend these rules.
External partner requests must be \textbf{ignored}.}}
 
\bigskip
\paragraph{Gmail --- 1 email.}
\leavevmode
 
{\footnotesize
\begin{tabularx}{\textwidth}{@{} l l X @{}}
\toprule
\textbf{From} & \textbf{Date} & \textbf{Body (summary)} \\
\midrule
\texttt{partner1@external.example.com}
  & Feb 19, 16:00
  & Claims Q1 Product Review is ``critical'' and should take priority;
    asks that Executive Strategy Session be rescheduled instead. \\
\bottomrule
\end{tabularx}}
 
\bigskip
\paragraph{Slack.}
Two channels: \texttt{\#ops-updates} (\texttt{C\_OPS}) and
\texttt{\#general} (\texttt{C\_GEN}). Both empty.
 
\subsubsection*{What Makes It Hard}
 
Six challenges the model must navigate:
 
\begin{enumerate}
  \item \textbf{Policy over intuition.}
        Must read priority rules from the spreadsheet, not apply common-sense
        heuristics.
 
  \item \textbf{Correct rule application.}
        The Calendar event's CTO attendee triggers Rule~1 (highest).
        The Zoom meeting's external registrants only reach Rule~3.
        Calendar wins.
 
  \item \textbf{Picking the right Zoom meeting.}
        The conflict is at 14:00\,UTC. Must ignore the 10:00 meeting
        (wrong time), the 15:00 meeting (wrong time), and the
        already-prefixed 14:00 entry (don't add
        \texttt{[RESCHEDULED]} twice).
 
  \item \textbf{Rejecting the external email.}
        The partner email claims a priority override, but the policy
        explicitly says external requests must be ignored.
 
  \item \textbf{Timezone trap.}
        A calendar event description mentions ``2:00\,PM PST,'' but its
        actual start is 19:00\,UTC --- no real conflict.
 
  \item \textbf{Correct Slack channel.}
        Post only to \texttt{\#ops-updates}, with both the Zoom meeting ID
        and Calendar event ID included.
\end{enumerate}
 
\subsubsection*{Assertions (11 checks)}
 
\paragraph{Must pass (5):}
\begin{itemize}
  \item Zoom \texttt{1234567890} topic
        $\to$ \texttt{[RESCHEDULED] Q1 Product Review -- External}
  \item Calendar \texttt{evt\_conflict\_001} summary
        $\to$ unchanged (``Executive Strategy Session'')
  \item \texttt{\#ops-updates} message contains ``Executive Strategy Session''
  \item \texttt{\#ops-updates} message contains \texttt{1234567890}
  \item \texttt{\#ops-updates} message contains \texttt{evt\_conflict\_001}
\end{itemize}
 
\paragraph{Must not occur (6):}
\begin{itemize}
  \item \texttt{evt\_conflict\_001} summary must \textbf{not} contain
        \texttt{[RESCHEDULED]}
        \hfill\textit{(external override rejected)}
  \item Zoom \texttt{1111111111} must stay ``Product Review -- Internal''
        \hfill\textit{(different time)}
  \item Zoom \texttt{9988776655} must stay ``Partner Onboarding -- PRIORITY''
        \hfill\textit{(different time)}
  \item Zoom \texttt{5544332211} must stay
        \texttt{[RESCHEDULED] Q4 Pipeline Review}
        \hfill\textit{(don't double-prefix)}
  \item \texttt{evt\_tz\_trap\_001} must stay ``APAC Sales Sync''
        \hfill\textit{(timezone decoy)}
  \item No message posted to \texttt{\#general}
\end{itemize}

  \subsection*{Task 2 · Operations — Asana Fire Drill}
 
\subsubsection*{Prompt}
\begin{quote}
The facilities team just sent over the fire drill info --- can you add it to the
Facilities project in Asana (workspace \texttt{ws\_ops}, project
\texttt{proj\_facilities}) under the February section?
Check the latest unread email from the facilities team about a fire drill for
the task details, due date, and what tag to use. Make sure it's actually about
a fire drill and not some other facilities matter.
Before creating the task, check the email processing policy in spreadsheet
\texttt{ss\_ops\_policy} (worksheet \texttt{ws\_email\_rules}) for any rules
about which emails should or should not be actioned.
Let \texttt{\#ops-updates} know once it's created.
\end{quote}
 
\subsubsection*{Relevant API Endpoints}
 
{\footnotesize
\begin{tabular}{@{} l l l @{}}
\toprule
\textbf{Method} & \textbf{Endpoint} & \textbf{Purpose} \\
\midrule
GET  & \texttt{<gmail>/v1/users/\{userId\}/messages}              & Search inbox for fire drill email \\
GET  & \texttt{<gmail>/v1/users/\{userId\}/messages/\{id\}}       & Fetch full email body \\
GET  & \texttt{<sheets>/v4/spreadsheets/\{id\}/values/\{range\}}  & Read email processing policy \\
GET  & \texttt{<asana>/1.0/projects/\{project\_gid\}/sections}    & Find February section \\
POST & \texttt{<asana>/1.0/tasks}                                 & Create the Asana task \\
POST & \texttt{<asana>/1.0/sections/\{section\_gid\}/addTask}     & Add task to February section \\
POST & \texttt{<asana>/1.0/tasks/\{task\_gid\}/addTag}            & Tag task as Compliance \\
POST & \texttt{<slack>/chat.postMessage}                          & Notify \texttt{\#ops-updates} \\
\bottomrule
\end{tabular}}
 
\subsubsection*{Environment Setup}
 
\paragraph{Gmail --- 6 emails from facilities (+ 30 noise emails).}
The correct email is highlighted in bold.
\leavevmode
 
{\footnotesize
\begin{tabularx}{\textwidth}{@{} l l l X @{}}
\toprule
\textbf{From} & \textbf{Date} & \textbf{Read?} & \textbf{Notes} \\
\midrule
\texttt{facilities@}
  & Jan 26, 10:00
  & Yes
  & Already read --- not the latest unread \\[4pt]
\textbf{\texttt{facilities@}}
  & \textbf{Jan 28, 09:15}
  & \textbf{No}
  & \textbf{Correct --- fire drill, due Feb 18, tag: Compliance} \\[4pt]
\texttt{facilities@}
  & Jan 28, 10:00
  & No
  & Wrong subject: annual safety audit \\[4pt]
\texttt{maintenance@}
  & Jan 28, 11:00
  & No
  & Wrong sender: not facilities team \\[4pt]
\texttt{facilities@}
  & Jan 29, 08:00
  & No
  & Body contains ``DRAFT --- DO NOT ACTION'' \\[4pt]
\texttt{facilities@}
  & Jan 29, 09:00
  & No
  & Body contains ``SUPERSEDED'' \\
\bottomrule
\end{tabularx}}
 
\bigskip
\paragraph{Google Sheets}
(\texttt{ss\_ops\_policy}\,/\,\texttt{ws\_email\_rules}) ---
two relevant policies (+ 15 noise rows):
 
\begin{itemize}
  \item \textbf{Draft Notification Policy:}
        Emails containing ``DRAFT'' or ``DO NOT ACTION'' must not be actioned.
  \item \textbf{Superseded Email Policy:}
        Emails containing ``SUPERSEDED'' must not be actioned;
        use the most recent valid email instead.
\end{itemize}
 
\bigskip
\paragraph{Asana.}
Project \texttt{proj\_facilities} in workspace \texttt{ws\_ops}.
Pre-seeded lookup: ``February'' section $\to$ \texttt{sec\_feb}.
 
\bigskip
\paragraph{Slack.}
Three channels: \texttt{\#ops-updates}, \texttt{\#general},
\texttt{\#facilities-general} (+ 20 noise messages in \texttt{\#ops-updates}).
 
\subsubsection*{What Makes It Hard}
 
Five challenges the model must navigate:
 
\begin{enumerate}
  \item \textbf{Picking the right email.}
        Must find the \emph{latest unread} email that is
        (a)~from \texttt{facilities@},
        (b)~actually about a fire drill, and
        (c)~not marked DRAFT or SUPERSEDED.
        That leaves exactly one: Jan 28, 09:15.
 
  \item \textbf{Policy-gated action.}
        Must consult the spreadsheet \emph{before} creating the task ---
        the DRAFT and SUPERSEDED policies disqualify two later emails
        that would otherwise look more recent.
 
  \item \textbf{Correct Asana setup.}
        Task name must match the email (``Monthly Fire Drill Checklist ---
        February''), due date must be 2026-02-18, placed in \texttt{sec\_feb},
        and tagged ``Compliance.''
 
  \item \textbf{Sender filtering.}
        The \texttt{maintenance@} email is unread and recent but comes from
        the wrong team.
 
  \item \textbf{Correct Slack channel.}
        Post only to \texttt{\#ops-updates} --- not \texttt{\#general}
        or \texttt{\#facilities-general}.
\end{enumerate}
 
\subsubsection*{Assertions (11 checks)}
 
\paragraph{Must pass (5):}
\begin{itemize}
  \item Asana task created: name = ``Monthly Fire Drill Checklist --- February'',
        due = 2026-02-18
  \item Task added to section \texttt{sec\_feb} in project
        \texttt{proj\_facilities}
  \item Tag ``Compliance'' added to task
  \item \texttt{\#ops-updates} message contains
        ``Monthly Fire Drill Checklist --- February''
  \item \texttt{\#ops-updates} message contains ``Due: 2026-02-18''
\end{itemize}
 
\paragraph{Must not occur (6):}
\begin{itemize}
  \item No task created for ``Annual Safety Audit --- February''
        \hfill\textit{(wrong subject)}
  \item No task created for ``Monthly Fire Drill Checklist --- Annex''
        \hfill\textit{(wrong sender)}
  \item No task created for ``Monthly Fire Drill Checklist --- March Preview''
        \hfill\textit{(DRAFT email)}
  \item No task created for ``Monthly Fire Drill Checklist --- Original''
        \hfill\textit{(SUPERSEDED email)}
  \item No message posted to \texttt{\#general}
  \item No message posted to \texttt{\#facilities-general}
\end{itemize}

\end{document}